\documentclass{article}

\usepackage[preprint]{neurips_2026}

\usepackage[utf8]{inputenc}
\usepackage[T1]{fontenc}
\usepackage{hyperref}
\usepackage{url}
\usepackage{booktabs}
\usepackage{amsfonts}
\usepackage{nicefrac}
\usepackage{microtype}
\usepackage{xcolor}
\usepackage{graphicx}
\usepackage{xcolor}
\usepackage{amsmath}
\usepackage{listings}
\definecolor{successgreen}{HTML}{2A9D8F}  % match the green in your figure
\definecolor{failureorange}{HTML}{E76F51} % match the orange in your figure
\definecolor{codecomment}{HTML}{6A737D}
\definecolor{codekeyword}{HTML}{D73A49}
\definecolor{codestring}{HTML}{032F62}
\lstdefinestyle{keystonepy}{
  language=Python,
  basicstyle=\ttfamily\footnotesize,
  keywordstyle=\color{codekeyword}\bfseries,
  commentstyle=\color{codecomment}\itshape,
  stringstyle=\color{codestring},
  showstringspaces=false,
  breaklines=true,
  columns=fullflexible,
  keepspaces=true,
  xleftmargin=0.5em,
  aboveskip=0.4em,
  belowskip=0.4em,
}
\newcommand{\success}{\textcolor{successgreen}{\text{success}}}
\newcommand{\failure}{\textcolor{failureorange}{\text{failure}}}
\newcommand{\green}[1]{\textcolor{successgreen}{#1}}

\newcommand{\orange}[1]{\textcolor{failureorange}{#1}}

\newcommand{\name}{KeyStone}

\title{Geometry Guided Self-Consistency for Physical AI}

\author{
  \textbf{Yinwei Dai}$
  \quad
  \textbf{Zhuofu Chen}$
  \quad
  \textbf{Lijie Yang}$
  \quad
  \textbf{Ravi Netravali}$
  \\
  Princeton University\\
    \texttt{\{yinweid,zhuofuc,ly322,rnetravali\}@princeton.edu}
}

\begin{document}

\maketitle

% !TeX root = ./paper.tex

\begin{abstract}

State-of-the-art physical AI models generate a chunk of actions per inference through diffusion or flow matching, iteratively refining an initial noise sample into an action trajectory. Because this inference process is inherently stochastic, committing to a single trajectory per round is brittle, and this brittleness compounds across the many sequential rounds that comprise a complete episode. We introduce \name{}, an inference-time self-consistency method for diffusion-based action generation that draws $K$ candidate action chunks in parallel from a shared model context, clusters them in continuous action space, and returns the medoid of the largest cluster -- no additional model required. Two properties make this practical. First, the compact nature of action trajectories makes diffusion inference memory-bandwidth bound, leaving spare compute capacity to run $K$ chains in parallel with no additional wall-clock latency. Second, unlike token or pixel spaces where distance carries no semantic meaning and selection requires a learned judge, action chunks are geometrically structured such that Euclidean distance directly reflects physical similarity, making selection principled and judge-free. Across diverse vision-language-action models (VLAs) and world-action models (WAMs), \name{} improves task success rates by up to \textbf{13.3\%} over single-trajectory sampling with negligible latency overhead, while having on par accuracy with model-based selectors at no training cost. We open source \name{} at \url{https://github.com/dywsjtu/keystone}.

\end{abstract}
% !TeX root = ./paper.tex

\section{Introduction}
\label{sec:introduction}

Diffusion and flow-matching models have become the dominant paradigm
for action generation in physical AI. They power
vision-language-action models (VLAs)~\citep{diffusion_policy,pi05,groot,smolvla,xvla}
as well as world-action models (WAM)~\citep{Fast-WAM,dream_zero,cosmos} that couple action generations with a future-observation prediction. Across this family, the action generation paradigm is the same: at each call, the model denoises an initial noise sample into a \emph{chunk} of actions, which is then executed open-loop on the robot before the model is queried again for the next chunk. This design captures the multimodal action distributions inherent in manipulation tasks and produces a smooth, temporally coherent control over continuous action spaces. A complete physical AI task, therefore, consists of many sequential rounds of action chunk sampling and execution.

\begin{figure}[t]
  \centering
  \includegraphics[width=\linewidth]{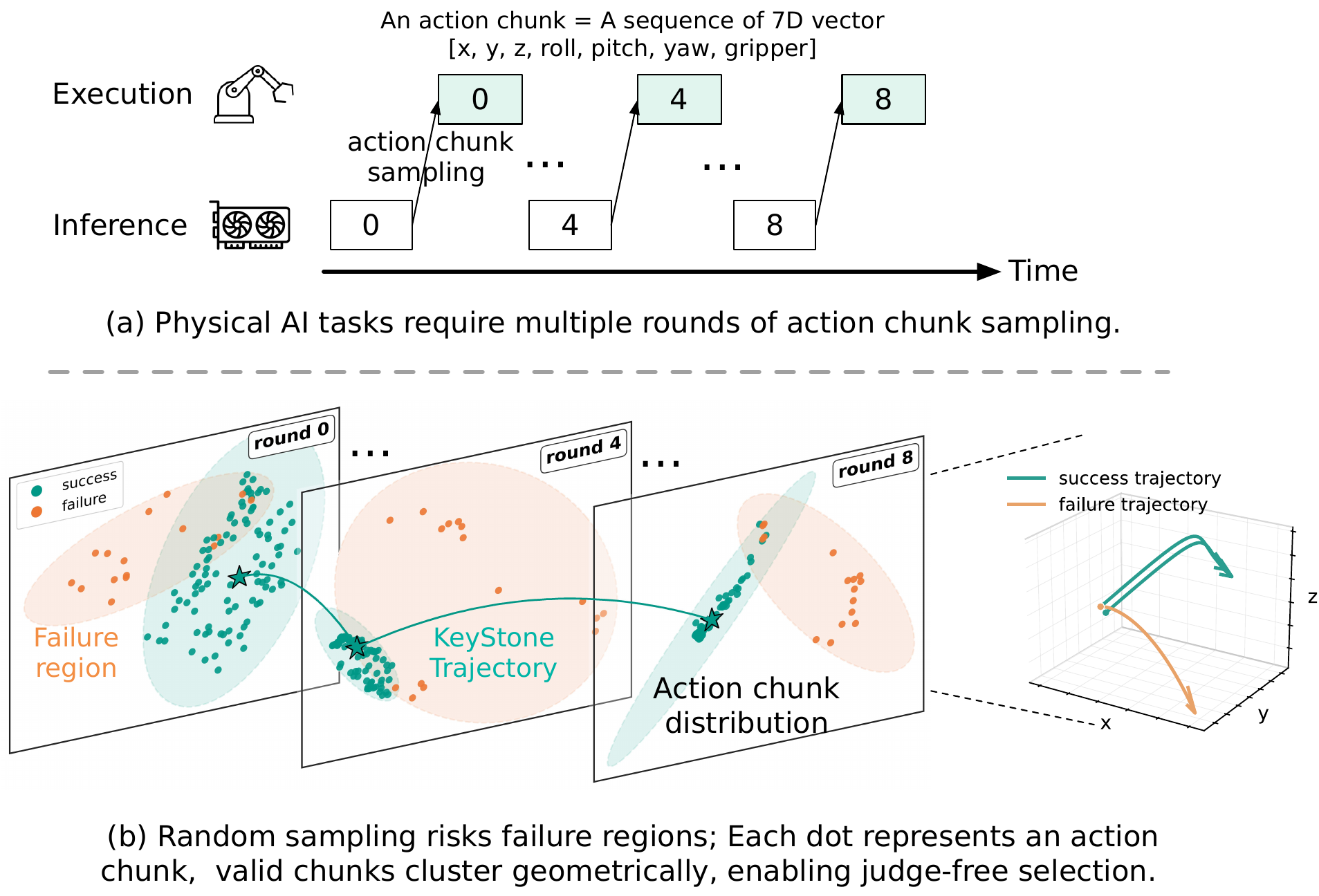}
  % \caption{%
  %   \textbf{Brittleness of single-trajectory sampling in physical AI.}
  %   \textbf{(a)}~A complete episode consists of many rounds of action
  %   chunk sampling and open-loop execution. Each round independently
  %   draws an action chunk from a stochastic diffusion or flow-matching
  %   process, so per-round sampling noise compounds across the episode.
  %   \textbf{(b)}~To probe per-round sampling behavior, we fix a known
  %   successful rollout from GR00T N1.6~\citep{groot} on
  %   SIMPLER~\citep{simpler} as the \emph{reference
  %   trajectory} (green stars). At each round $k$, we then sample many
  %   alternative action chunks from the same model conditioned on the
  %   reference's observation history at round $k$, execute each
  %   candidate [CONTINUATION], and label the resulting episode as
  %   success (green) or failure (orange). Each chunk is projected to
  %   2D via PCA fit per round. Successful chunks cluster densely while
  %   failures occupy sparse, scattered regions, making the densest
  %   cluster a natural target for consensus selection in continuous
  %   action space.
  % }
  \caption{%
  \textbf{Brittleness of single-trajectory sampling in physical AI.}
  \textbf{(a)}~A complete episode consists of many rounds of action
  chunk sampling and open-loop execution. Each round's stochastic sampling can derail the task.
  \textbf{(b)}~At each round $k$, we draw many candidate chunks from
  GR00T N1.6~\citep{groot} on SIMPLER~\citep{simpler}, execute each to
  episode end, and label the outcome \success{} or \failure{} (chunks
  are projected via per-round PCA). \success{} chunks cluster densely
  while \failure{} ones scatter. The inset shows the corresponding
  end-effector paths in 3D: \success{} chunks lead to closer 3D trajectories that reach the goal,
  while a \failure{} one clearly diverges. KeyStone exploits this by sampling multiple action chunks and selecting
  the medoid of the largest cluster at each round (\green{green stars});
  chained together, these selections form the \emph{KeyStone
  trajectory}.
  % \textbf{(b)}~To probe per-round sampling behavior, we start from a successful rollout of GR00T N1.6~\citep{groot} on SIMPLER~\citep{simpler} (the \emph{reference trajectory}, \green{green stars}), replay it exactly through round $k{-}1$, and resample alternative chunks at round $k$. We execute each alternative, continue the rollout to episode end, and label the outcome \success{} or \failure{}; chunks
  % are projected via per-round PCA. Even with identical preceding
  % rounds, alternatives at round $k$ routinely fall into failure
  % regions---yet \green{successes} cluster densely while
  % \orange{failures} scatter, suggesting the densest cluster as a
  % natural target for consensus selection.
  }
  \label{fig:keystone}
\end{figure}

This sampling-and-execution loop, however, is inherently stochastic.
At each round, the policy starts from a fresh noise sample and
iteratively refines it into an action chunk; different initial noises
lead to different chunks, and not all of them succeed. Some fall
into \emph{failure regions} of the action space---chunks that, when
executed, derail the task even when prior rounds were successful
(Fig.~\ref{fig:keystone}b). Worse, this brittleness compounds: an
episode comprises many rounds of chunk sampling
(Fig.~\ref{fig:keystone}a), so even a small per-round probability of
landing in a failure region accumulates into a substantial
episode-level failure rate. Committing to a single sample per round,
the standard practice today, leaves task success at the mercy of
sampling noise.

A natural response to sampling stochasticity is test-time scaling:
draw multiple candidates and select (e.g., self-consistency~\citep{self-consistency}), as studied extensively in language models~\citep{self-consistency,math-verifier,prm}. Recent work brings this paradigm to physical AI~\citep{robomonkey,taco}, sampling multiple action candidates and selecting among them with a learned selector trained on the physical AI model's fine-tuning dataset to score how well a candidate matches the in-distribution success modes. This adds non-trivial costs: an extra model to train per task and embodiment (hundreds of H100 hours in TACO~\cite{taco} for $\pi_{0.5}$ on LIBERO~\cite{libero}, which exceeds $\pi_{0.5}$'s fine-tuning itself), plus an extra forward pass per round at inference.

%\textbf{Our key observation is that this learned selector is redundant: the very distribution the selector is trained to estimate is already baked into the model and surfaces directly in the geometry of sampled action chunks. \dy{Why do we tightly tie our observation to the redundancy of learned selector? We will be questioned why not to compare our approach with learned selector approaches empirically to show the redundancy. Can't we directly frame it as we find exploiting the geometry of sampled action chunks is enough to boost the accuracy with zero overhead.}} \rn{Our key observation is that a learned selector is unnecessary: the geometry of sampled action chunks directly encodes the model's concentration on in-distribution behaviors, making it sufficient for judge-free, accurate selection.} \rn{Our key observation is that a learned selector is unnecessary: the very distribution it would estimate is already baked into the model and surfaces directly in the geometry of sampled action chunks.}
\textbf{Our key observation is that the geometry of sampled action chunks is surprisingly powerful: it directly encodes the model's concentration on in-distribution behaviors, making the overheads associated with explicit learned selectors largely unwarranted.} Action chunks live in a continuous space whose structure is constrained -- every action is bounded by the robot's physical degrees of freedom -- so the Euclidean distance between two action chunks directly reflects whether they would drive the robot to similar execution trajectories. Unlike token or pixel spaces where Euclidean distance carries no semantic meaning (necessitating judges to bridge the gap), action spaces are physically constrained and low-dimensional, making distance a direct proxy for behavioral similarity. The model's concentration on in-distribution behaviors therefore manifests directly as a dense cluster of geometrically proximate candidates; Fig.~\ref{fig:keystone}b shows this structure empirically.

Capitalizing on these insights, we present \textbf{\name{}}, an inference-time self-consistency method that, at each round, draws $K$ candidate action chunks in parallel from a single shared context, clusters them in the continuous action space, and returns the medoid of the largest cluster as the chunk to execute. Importantly, because diffusion inference for action generation is memory-bandwidth bound, \name{}'s parallel sampling fits within spare compute capacity and adds no additional wall-clock latency per round. This is because, unlike image and video diffusion, where the diffusion processes are compute-bound due to the large input dimensions, action diffusion operates over compact trajectories (typically $O(100)$ dimensions), shifting the bottleneck from compute to memory bandwidth and leaving substantial parallel compute headroom unused.

We evaluate across diverse VLAs and WAMs, and workloads, setting $K$ to the largest value that preserves per-round latency, and find that \name{} improves task success rates by up to \textbf{13.3\%} over single-trajectory sampling and is comparable with the model-based test-time scaling approach TACO at zero training cost. More broadly, \name{} shows that physical AI models inherently possess everything needed for successful self-consistency: the geometry to select without a judge, and the compute headroom to sample without added latency.

% !TeX root = ./paper.tex

\section{Background and Related Work}
\label{sec:background}

%\paragraph{Action generation via diffusion and flow matching.}
%Physical AI models map observational inputs---camera images, robot states, task instructions---to actions, low-level motor commands such as gripper signals or joint positions. To meet real-time constraints, state-of-the-art models output \emph{action chunks}---short sequences of $N$ future actions---generated by iteratively refining an initial noise sample. This paradigm underlies diffusion- and flow-matching-based VLAs such as $\pi_0$~\citep{pi05}, GR00T~\citep{groot}, and SmolVLA~\citep{smolvla}, as well as world-action models such as FastWAM~\citep{Fast-WAM}, Cosmos~\citep{cosmos}, and DreamZero~\citep{dream_zero}, which additionally predict future observations (e.g., video frames). At deployment, a complete task consists of many sequential rounds of action chunk generation and execution.

\paragraph{Action generation via diffusion and flow matching.} Physical AI models map observational inputs---camera images, robot states, task instructions---to actions, low-level motor commands such as gripper signals or joint positions. To meet real-time constraints, state-of-the-art models output \emph{action chunks}, or short sequences of future actions generated by iteratively refining an initial noise sample via diffusion or flow matching. This paradigm underlies VLAs such as $\pi_0$~\citep{pi05}, GR00T~\citep{groot}, and SmolVLA~\citep{smolvla}, as well as world-action models (WAMs) such as FastWAM~\citep{Fast-WAM}, Cosmos~\citep{cosmos}, and DreamZero~\citep{dream_zero}, which additionally predict future observations such as video frames.

\paragraph{Test-time scaling in text and image generation.}
Sampling multiple candidates and selecting among them is a
well-established way to improve generative model outputs.
\emph{Model-free} approaches avoid any auxiliary scorer: they
select among candidates using a property of the candidates
themselves, such as exact-match agreement on discrete final
answers~\citep{self-consistency} or the model's own output confidence~\citep{deepconf}. \emph{Model-based} approaches train a separate scorer when no such property is available---verifiers and process
reward models~\citep{math-verifier,prm} for reasoning, and for image generation~\citep{image-reward,pick-a-pic,inferencetimescalingdiffusionmodels}.

\paragraph{Test-time scaling in action generation.} Earlier work extended test-time scaling to autoregressive VLAs. RoboMonkey~\citep{robomonkey} 
samples $K$ action candidates and selects with a VLM-based verifier; 
V-GPS~\citep{vgps} trains a value function for the same purpose; 
MG-Select~\citep{mgselect} scores candidates model-free via KL divergence from a condition-masked reference distribution.
As physical AI has shifted toward diffusion- and flow-matching architectures, recent work has adapted selection to this paradigm. 
TACO~\citep{taco} trains a pseudo-count estimator on the finetuning data and selects the candidate with the highest estimated count, constraining outputs to the finetuning data distribution. However, the auxiliary estimator adds inference overhead and must be retrained from the finetuning data and model-internal features at a cost of \textbf{hundreds of H100 hours} per task or embodiment---prohibitive given how often physical AI models are finetuned.
The existing model-free direction does not transfer either: methods such as MG-Select rely on token-level confidence signals that an autoregressive model exposes natively from its final classifier, which diffusion- and flow-matching models lack. KDPE~\citep{KDPE} applies kernel density estimation to the \textbf{final action} of $N$ independent action chunks without a learned model. However, this does not scale to the dimensionality of full action chunks, forcing selection based on a single action point and discarding geometric signal across the rest of the chunk. KDPE also requires a task-specific manifold-aware kernel for each action representation. \name{} instead exploits the general geometric structure of continuous action spaces to cluster over complete action chunks without task specific design, capturing richer signal across the full chunk. Overall, \name{} is, to our knowledge, the first latency-neutral, judge-free test-time scaling method for diffusion- and flow-matching VLAs and WAMs, grounded in the geometric structure of continuous action spaces. %Overall, \name{} is, to our knowledge, the first method to identify and exploit this geometric structure, enabling latency-neutral, judge-free test-time scaling across diffusion- and flow-matching VLAs and WAMs.

%\paragraph{Test-time scaling in action generation.} Earlier work extended test-time scaling above to autoregressive VLAs, which output actions as discrete tokens. RoboMonkey~\citep{robomonkey} samples $K$ action candidates and selects with a VLM-based verifier; V-GPS~\citep{vgps} trains a value function for the same purpose; MG-Select~\citep{mgselect} scores candidates model-free via KL divergence from a condition-masked reference distribution. As physical AI has shifted toward diffusion- and flow-matching architectures, recent work has followed by adapting model-based selection to this paradigm: TACO~\citep{taco} trains a pseudo-count estimator on the finetuning data and selects the candidate with the highest estimated count, constraining outputs to the finetuning data distribution. However, the auxiliary model incurs additional inference overhead and must be retrained whenever the underlying model is finetuned for a new embodiment or task, which is common in physical AI deployment. The existing model-free direction does not transfer either: methods such as MG-Select rely on token-level confidence signals that an autoregressive model exposes natively from its final classifier, which diffusion- and flow-matching models lack. \name{} is, to our knowledge, the first model-free test-time scaling method for diffusion- and flow-matching VLAs and WAMs, exploiting the natural geometric structure of action chunks themselves.

% !TeX root = ./paper.tex
\section{\name}
\label{sec:system}

We introduce \name{}, a drop-in inference-time wrapper for stochastic
diffusion- and flow-matching physical AI models. At each round of the standard sample-and-execute loop,
\name{} draws multiple action chunks from the same encoded observations
(e.g., camera images and robot states) and task instruction, identifies
the largest cluster in continuous action space, and executes that
cluster's medoid -- a real sampled chunk rather than a synthesized
average. The method is therefore model-free: it uses only the geometry of
the sampled action chunks, rather than a learned or task-specific
selector.

\subsection{Method Overview}
\label{sec:method-overview}

Let $c_t$ denote the encoded context at round $t$, including the task
instruction, camera observations, and robot state. A diffusion- or
flow-matching model maps this context and an initial noise tensor
$\epsilon$ to an action chunk:
\begin{equation}
  a = G_\theta(c_t, \epsilon),
  \qquad \epsilon \sim \mathcal{N}(0, I),
\end{equation}
where $a$ is the sequence of future actions that will be executed
sequentially before the next model call. Standard deployment draws one
$\epsilon$ and commits to the resulting chunk. \name{} instead draws
$K$ independent noises
$\{\epsilon_i\}_{i=1}^K$ under the same context and obtains
\begin{equation}
  a_i = G_\theta(c_t, \epsilon_i),
  \qquad i=1,\dots,K.
\end{equation}
It then chooses one candidate $a^\star \in \{a_i\}_{i=1}^K$ to execute.
We return an actual generated action chunk, rather than an averaged trajectory,
so the output remains on the model's sampled action manifold and does
not interpolate between distinct modes.

\subsection{Latency-Preserving Parallel Sampling }
\label{sec:parallel-candidates}

The key efficiency insight is that $K$ parallel chains can add no
wall-clock latency when the dominant inference cost is context
computation, because that context is shared across all candidates.
Action diffusion operates over compact trajectories rather than large
image or video latents, so the denoising computation is small and
memory-bandwidth bound. This leaves free compute headroom for parallel sampling: once the
shared prefix has been computed, adding candidates mainly adds small
action-state tensors and batched denoising or flow updates.

Candidates share the same conditioning context. For VLAs, the
image and language inputs are identical across candidates, so \name{}
computes the vision language model prefix once and reuses the resulting
features or the transformer key-value (KV) cache for all $K$ denoising
chains. For WAMs, the same idea applies. The observation is passed through the video generation backbone once to produce the shared latent world embeddings. In implementation, this is a
logical batch expansion: shared prefix feature tensors are first given a
candidate dimension and expanded with \texttt{torch.expand}, so the $K$
candidates share the same underlying storage rather than materializing
$K$ copies of the context. At each denoising or flow update step,
the model processes those $K$ candidates as a single batched forward
pass. In practice, we choose $K$ by profiling the target model on the
deployment hardware and use the largest value whose per-round latency
matches single-sample inference. We discuss the implementation details in the appendix.

\subsection{Candidate Selection}
\label{sec:candidate-selection}

The core insight behind our selector is that action chunks are ordered
sequences of bounded robot commands. Nearby chunks in continuous action
space correspond to physically similar execution trajectories by construction,
so the largest dense region provides a natural model-free notion of
consensus. \name{} therefore clusters the sampled chunks in action space
and executes the medoid rather than the centroid of the dominant
cluster. This guarantees that the selected action is a real model
output, avoiding interpolation between distinct modes that could produce
trajectories that are physically invalid and inconsistent with the
model's output distribution.

To achieve that, each candidate chunk is first flattened into a vector
$x_i$, and all pairwise L2 distances are computed in one batched tensor
operation:
\begin{equation}
  \Delta_{ij} = \lVert x_i - x_j \rVert_2 .
\end{equation}

Before clustering, \name{} first checks whether clustering is needed.
This guard handles a simple failure mode of forced clustering: when the
candidate distribution is already unimodal, k-means can split one valid
cluster into arbitrary halves, causing the selector to return the
medoid of an artificial subcluster rather than the global medoid. To
avoid this, \name{} computes the global medoid
$m = \arg\min_i \sum_j \Delta_{ij}$ and measures how far the sample
mean $\bar{x}$ lies from that medoid, normalized by
the median pairwise distance:
\begin{equation}
  s =
  \frac{\lVert \bar{x} - x_m \rVert_2}
       {\operatorname{median}_{i<j}\Delta_{ij} + \epsilon}.
\end{equation}
If $s$ is below a threshold $\tau$ (default $0.3$), the candidates are
treated as a single compact cluster and \name{} returns the global
medoid $a_m$. The threshold is intentionally conservative: it only
suppresses clustering when the mean and medoid nearly agree relative to
the typical candidate spread, which is the regime where a cluster split
is least informative. Otherwise, \name{} runs cluster-medoid selection.
The number of clusters is exposed as a hyperparameter $C$; we use
$C=2$ by default because most manipulation decisions have one dominant
valid mode and scattered failure samples, so the main decision is
whether a candidate belongs to the consensus mode or not. Given $C$,
\name{} runs a small k-means loop over the flattened candidate chunks.
We initialize the centroids by randomly selecting $C$ candidates from
the sampled batch. We iterate up to 10 times and stop early when the
cluster assignments do not change between successive iterations. This
assigns each candidate to a cluster,
\begin{equation}
  z_i \in \{1,\dots,C\},
  \qquad
  \mathcal{C}_c = \{i : z_i = c\}.
\end{equation}
It then chooses the largest cluster and returns that cluster's medoid:
\begin{equation}
  c^\star = \arg\max_{c \in \{1,\dots,C\}} |\mathcal{C}_c|,
  \qquad
  \mathcal{C}^\star = \mathcal{C}_{c^\star},
  \qquad
  i^\star =
  \arg\min_{i \in \mathcal{C}^\star}
  \sum_{j \in \mathcal{C}^\star} \Delta_{ij},
  \qquad
  a^\star = a_{i^\star}.
\end{equation}
The selected chunk is thus both representative of the dominant sampled
mode and \textbf{guaranteed to be one of the model's own outputs}. The rule has
no trained parameters, no auxiliary scorer. Its cost is small for the sample counts used in our
evaluations (<16): the filter uses the pairwise distances already computed for
medoid selection, k-means runs over only $K$ compact action chunks, and
the within-cluster medoid reuses the same pair-wise distance matrix.

\subsection{Deployment Loop}
\label{sec:deployment-loop}

At deployment, \name{} is a lightweight wrapper around the original
policy sampler. At each round it intercepts the generation call,
runs the parallel sampling and selection procedures of
Sections~\ref{sec:parallel-candidates}--\ref{sec:candidate-selection},
and returns a single action chunk to the downstream robotic application in the
original output format. The selection step itself adds negligible
overhead: its pairwise distances, k-means, and medoid arg-mins are
all batched tensor operations over only $K$ compact action chunks.
We describe the implementation details in the appendix.

% !TeX root = ./paper.tex

\section{Evaluation}
\label{sec:evaluation}

We evaluate \name{} along the two axes implied by its design:
whether parallel candidate sampling preserves wall-clock latency,
and whether \name{}'s candidate selection improves task success
over the single-sample baseline while matching model-based test-time scaling approach TACO. 

\subsection{Experimental Setup}
\label{sec:experimental-setup}

\paragraph{Testbed.}
We implement \name{} as an inference wrapper across
three evaluation testbeds: LeRobot~\citep{lerobot}, vla-eval~\citep{vla-eval},
and Fast-WAM~\citep{Fast-WAM}. In each testbed, the $K=1$ baseline is
the original model evaluated under the configuration used to reproduce
its reported behavior, including the benchmark suite, action chunk size,
action post-processing, and per-task repeat episode count. For vla-eval, we use its benchmark Docker images so the simulator, dependencies,
and other setups exactly match the original baseline setup. The
baseline path is unchanged when the wrapper is disabled or when $K=1$.
For $K>1$, the wrapper intercepts action generation, constructs a
candidate batch from independent noise samples. The selector returns
one selected action chunk, and the testbed passes that chunk through the same downstream execution path used by the baseline. We run our evaluations on NVIDIA A6000-48GB GPUs.

\paragraph{Models.}
Across these three testbeds, we evaluate \name{} on physical AI models
spanning VLAs and WAMs, including $\pi_{0.5}$~\citep{pi05},
SmolVLA~\citep{smolvla}, GR00T N1.6~\citep{groot},
X-VLA~\citep{xvla}, StarVLA (Qwen VLM + GR00T action diffuser)~\citep{starvla}, and
Fast-WAM~\citep{Fast-WAM}.

\paragraph{Workloads.}
We use simulated manipulation workloads from LIBERO~\citep{libero}, SimplerEnv~\citep{simpler}, and RoboTwin
2.0~\citep{robotwin2}. For SimplerEnv, we evaluate both supported robot embodiments used
in our results: WidowX and Google Robot. The workloads span single-arm tabletop
manipulation, long-horizon instruction following, cross-embodiment
settings, and dual-arm manipulation.

\paragraph{Metrics.}
Our primary metric is task success rate. Within each workload, every
task is evaluated over the same number of episodes used by the
baseline configuration in their papers or implementations, typically 24--100 episodes per task mentioned above. Each
episode is one independent closed-loop rollout: the testbed
instantiates the simulator from one of the task's initial conditions
(object poses, target configurations, scene seeds), executes the
model in closed loop until the task succeeds or a fixed step budget
is reached, and records a binary success label using the task's
success criterion. The per-task success rate is the fraction of these
episodes that succeed, and the workload-level success rate reported
in our tables is the average of the per-task rates
within the workload. For each model--workload pair, we repeat this
full evaluation five times unless otherwise noted---each repeat
re-samples the per-episode sampling noise---and report the mean and
one-standard-deviation range across the five repeats. The reported
standard deviation therefore captures only the \emph{across-repeat
variation}; within-repeat noise is already absorbed into each repeat's
per-task success rate. We also report end-to-end per-round latency,
which includes model inference for sampling the $K$ candidates and
final selection, and peak GPU memory during the same
action-generation round.

\subsection{Latency and Overhead}
\label{sec:latency-overhead}

\begin{figure}[!htbp]
  \centering
  \includegraphics[width=\linewidth]{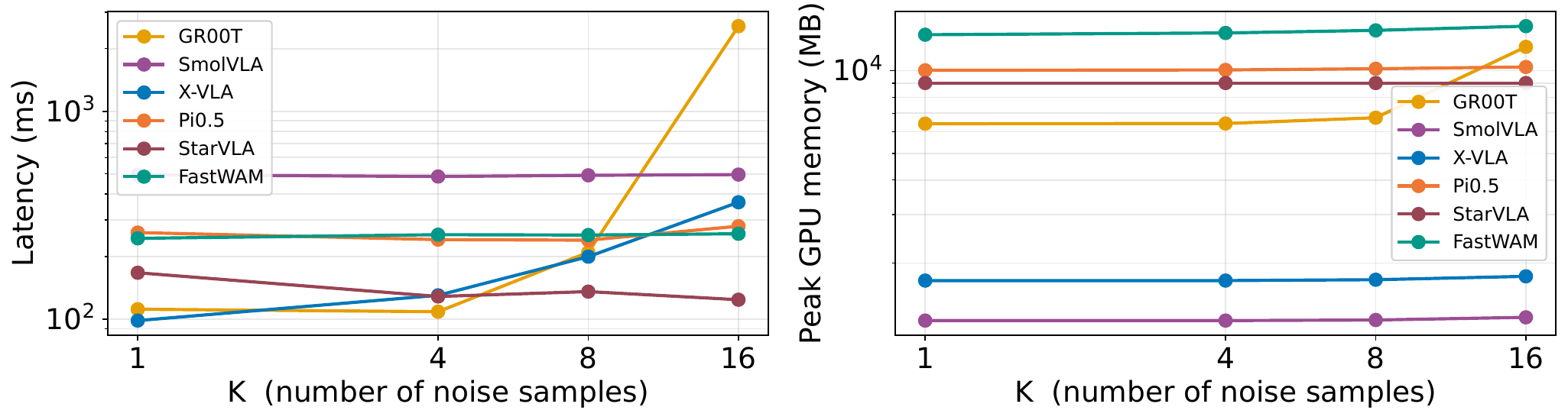}
  \caption{%
  \textbf{End-to-end per-round latency and peak GPU memory as the number of
  sampled action chunks increases.} We measure the wall-clock latency of one
  action-generation round at $K \in \{1,4,8,16\}$, including model
  inference, the overhead of sampling $K$ candidates, and \name{}'s
  selection step. We also report the peak GPU memory used by the same
  generation round.}
  \label{fig:latency}
\end{figure}

Figure~\ref{fig:latency} reports \name{}'s end-to-end latency and peak
GPU memory as the number of sampled candidates $K$ increases. The latency
measurement includes model inference, parallel candidate sampling, and the
cluster-based selection step. Drawing multiple candidate chunks in parallel
adds little latency over the operating range used in our evaluations. Peak GPU
memory also remains stable across these $K$ values.
This scaling comes from two properties. First, the shared model context is
encoded and stored in memory once per round rather than once per candidate.
Second, the per-candidate diffusion computation operates on compact action
chunks. This compact action representation makes the diffusion process memory-bandwidth bound,
leaving compute headroom for parallel candidate sampling without increasing latency. 
It also keeps the added activations and candidate-selection state small, adding little GPU memory overhead. 

We therefore choose the largest $K$ whose end-to-end latency
remains close to the $K=1$ baseline. For GR00T and X-VLA, latency begins to
inflate beyond $K=4$, so we use $K=4$ in the subsequent evaluation. For the
remaining models, we set $K=16$ as the latency stays close to the baseline.
These measurements assume the common single-robot deployment setting
with a dedicated GPU; Section~\ref{sec:limitations} discusses the
shared-accelerator case where the spare compute headroom that
\name{}'s parallel sampling relies on may shrink, and $K$ should be
re-profiled against the actual deployment workload.

\subsection{Task Success Rates}
\label{sec:main-results}

Table~\ref{tab:main-results} reports the workload-level success
rate---computed as described in
Section~\ref{sec:experimental-setup}---for every model--workload
pair, comparing the $K{=}1$ baseline against \name{} at the
per-model $K$ chosen in Section~\ref{sec:latency-overhead} ($K{=}4$
for GR00T~N1.6 and X-VLA, $K{=}16$ for the rest). Bold marks the
higher mean per row, and the $\Delta$ column gives the absolute
success rate improvement of \name{} over the baseline.

\begin{table}[t]
  \centering
  \small
  \setlength{\tabcolsep}{6pt}
  \renewcommand{\arraystretch}{1.1}
  \caption{\textbf{Aggregate task success rates (\%) on simulated
    manipulation workloads.} Mean $\pm$ one standard deviation across 5 runs, 
    where each run is macro-averaged across the tasks within each workload.
    $\Delta$ is the absolute (percentage-point) improvement of
    \name{} over the $K{=}1$ baseline.}
  \label{tab:main-results}
  \begin{tabular}{llcccc}
    \toprule
    \textbf{Workload} & \textbf{Model} & \textbf{$K$} &
    \textbf{Baseline ($K{=}1$)} & \textbf{\name{}} & \textbf{$\Delta$} \\
    \midrule
    LIBERO                  & $\pi_{0.5}$  & 16 & 96.8 $\pm$ 0.7 & \textbf{97.8 $\pm$ 1.1} & \green{\textbf{$+1.0$}} \\
    LIBERO                  & SmolVLA      & 16 & 50.4 $\pm$ 2.1 & \textbf{57.2 $\pm$ 1.3} & \green{\textbf{$+6.8$}} \\
    SimplerEnv -- WidowX    & GR00T~N1.6   & 4  & 50.0 $\pm$ 4.3 & \textbf{63.3 $\pm$ 2.9} & \green{\textbf{$+13.3$}} \\
    SimplerEnv -- WidowX    & X-VLA        & 4  & 92.7 $\pm$ 1.0 & \textbf{95.8 $\pm$ 1.0} & \green{\textbf{$+3.1$}} \\
    SimplerEnv -- WidowX    & StarVLA      & 16 & 52.8 $\pm$ 1.6 & \textbf{59.7 $\pm$ 1.3} & \green{\textbf{$+6.9$}} \\
    SimplerEnv -- Google Robot & GR00T~N1.6 & 4 & 79.4 $\pm$ 3.5 & \textbf{86.7 $\pm$ 1.5} & \green{\textbf{$+7.3$}} \\
    Robotwin 2.0            & Fast-WAM      & 16 & 90.0 $\pm$ 1.5 & \textbf{93.0 $\pm$ 1.0} & \green{\textbf{$+3.0$}} \\
    \bottomrule
  \end{tabular}
\end{table}

\name{} improves every model--workload pair, with absolute gains of
$+1.0$ to $+13.3$ percentage points over the single-sample baseline.
The improvements hold across both VLAs and the WAM, and across
LIBERO, SimplerEnv on WidowX and Google Robot, and RoboTwin 2.0.
The improvements appear on two axes---a higher mean
success rate and a smaller across-repeat standard deviation---both
arising from \name{}'s better per-round sampling. We describe
each effect in turn, then characterize the (model, workload) regime
in which the gains concentrate.

\paragraph{Mean improvement.}
At each action-generation round, \name{} selects the action chunk
from the dominant sampled cluster rather than committing to a
single stochastic draw, replacing the per-round failure rate of a
generic random sample with that of the model's dominant sampled mode.
Because each rollout consists of many rounds, this per-round
reduction compounds into the task-level success-rate gains in
Table~\ref{tab:main-results}. Every row improves; the largest gain
is $+13.3$ for GR00T~N1.6 on SimplerEnv--WidowX
($50.0\% \to 63.3\%$). The gain on Fast-WAM (+3.0) confirms that \name{}'s geometric selection generalizes beyond VLAs to WAMs, where the same geometric structure holds.

\paragraph{Variance reduction.}
The same mechanism also reduces the across-repeat standard
deviation of the success rate. \name{} consistently selects from the
dominant cluster across rounds, so successive evaluations yield more similar
task-level outcomes. The standard deviation drops from $4.3$ to
$2.9$ for GR00T~N1.6 on SimplerEnv--WidowX, from $3.5$ to $1.5$ on
SimplerEnv--Google Robot, from $2.1$ to $1.3$ for SmolVLA on LIBERO,
and from $1.6$ to $1.3$ for StarVLA on SimplerEnv--WidowX.

\paragraph{Region of benefit: high-variance baselines.}
\name{} delivers its largest gains on workloads where the $K{=}1$
baseline shows visibly high across-repeat variance, the direct
signature of a model whose single-sample inference is suffering
from sampling randomness. The four rows with the highest
$K{=}1$ standard deviations are exactly the four with the largest
mean gains: GR00T~N1.6 on SimplerEnv--WidowX
($\pm 4.3$, $\Delta\!=\!+13.3$), GR00T~N1.6 on
SimplerEnv--Google Robot ($\pm 3.5$, $\Delta\!=\!+7.3$), SmolVLA
on LIBERO ($\pm 2.1$, $\Delta\!=\!+6.8$), and StarVLA on
SimplerEnv--WidowX ($\pm 1.6$, $\Delta\!=\!+6.9$). All four also
receive a meaningful variance reduction. The mechanism is
consistent: workloads where single-sample inference visibly
scatters across repeats are exactly the workloads where \name{}
has the most scattered failure samples to suppress, so suppressing
them lifts the mean and tightens the spread together.

\begin{table}[!htbp]
  \centering
  \small
  \setlength{\tabcolsep}{6pt}
  \renewcommand{\arraystretch}{1.05}
  \caption{\textbf{Breakdown on LIBERO with $\pi_{0.5}$
    ($K{=}16$).} Every suite (each has 10 tasks) is already saturated at $K{=}1$, so
    \name{} contributes only small deltas that average
    to a $+1.0$ overall gain.}
  \label{tab:breakdown-libero-pi05}
  \begin{tabular}{lccc}
    \toprule
    \textbf{Suite} & \textbf{Baseline ($K{=}1$)} & \textbf{\name{}} & \textbf{$\Delta$} \\
    \midrule
    Spatial & 96.0 $\pm$ 1.4 & \textbf{98.0 $\pm$ 1.4} & \green{\textbf{$+2.0$}} \\
    Object  & 99.0 $\pm$ 1.4 & \textbf{99.5 $\pm$ 0.7} & \green{\textbf{$+0.5$}} \\
    Goal    & \textbf{98.0 $\pm$ 1.4} & 97.5 $\pm$ 0.7 & \orange{\textbf{$-0.5$}} \\
    Long    & 94.0 $\pm$ 1.4 & \textbf{95.0 $\pm$ 1.4} & \green{\textbf{$+1.0$}} \\
    \midrule
    Overall  & 96.8 $\pm$ 0.7 & \textbf{97.8 $\pm$ 1.1} & \green{\textbf{$+1.0$}} \\
    \bottomrule
  \end{tabular}
\end{table}

\begin{table}[!htbp]
  \centering
  \small
  \setlength{\tabcolsep}{6pt}
  \renewcommand{\arraystretch}{1.05}
  \caption{\textbf{Breakdown on SimplerEnv--WidowX with
    X-VLA ($K{=}4$).} Two tasks are already saturated at $K{=}1$
    ($\geq 98\%$); the overall $+3.1$ gain comes mostly from
    the improvement on the two less-saturated tasks.}
  \label{tab:breakdown-xvla-widowx}
  \begin{tabular}{lccc}
    \toprule
    \textbf{Task} & \textbf{Baseline ($K{=}1$)} & \textbf{\name{}} & \textbf{$\Delta$} \\
    \midrule
    Put Carrot on Plate     & \textbf{100.0 $\pm$ 0.0} & \textbf{100.0 $\pm$ 0.0} & $\phantom{+}0.0$ \\
    Put Eggplant in Basket  & 88.9 $\pm$ 6.4 & \textbf{95.8 $\pm$ 4.2} & \green{\textbf{$+6.9$}} \\
    Put Spoon on Tablecloth & 98.6 $\pm$ 2.4 & \textbf{100.0 $\pm$ 0.0} & \green{\textbf{$+1.4$}} \\
    Stack Green Cube        & 83.3 $\pm$ 4.2 & \textbf{87.5 $\pm$ 1.6} & \green{\textbf{$+4.2$}} \\
    \midrule
    Overall                  & 92.7 $\pm$ 1.0 & \textbf{95.8 $\pm$ 1.0} & \green{\textbf{$+3.1$}} \\
    \bottomrule
  \end{tabular}
\end{table}

\paragraph{Where the gains plateau.}
At the opposite end, the rows where the $K{=}1$ baseline already
operates with low variance and near-perfect accuracy give \name{}
no scattered failure mass to remove. $\pi_{0.5}$ on LIBERO, X-VLA
on SimplerEnv--WidowX, and Fast-WAM on RoboTwin~2.0 all sit above
$90\%$ at $K{=}1$ with $K{=}1$ standard deviations of $1.5$ or
less, and the corresponding gains are modest ($+1.0$, $+3.1$,
$+3.0$). Table~\ref{tab:breakdown-libero-pi05} drills into the
$\pi_{0.5}$ on LIBERO case: every one of the four LIBERO suites (each has 10 tasks) is
already at or above $94\%$ at $K{=}1$, so \name{} contributes only
small per-suite deltas (ranging from $-0.5$ to $+2.0$) that
average to a $+1.0$ macro gain.
Table~\ref{tab:breakdown-xvla-widowx} shows the X-VLA on
SimplerEnv--WidowX case, which is more heterogeneous: two of the
four tasks are already at or above $98\%$ at $K{=}1$ and contribute
essentially nothing, while the two less-saturated tasks (Eggplant
in Basket at $88.9\%$, Stack Green Cube at $83.3\%$) improve by
$6.9\%$ and $4.2\%$ respectively, averaging to a $3.1\%$ macro gain. This regime is where \name{}'s
benefit asymptotes: saturated baselines with high success rates and low variance directly implies that the model is already concentrating on the dominant
mode, leaving consensus selection nothing distinct to recover.
Symmetrically, \name{}'s benefit also vanishes at the opposite end
of the spectrum---models where the majority of samples are confidently wrong, in which case cluster-medoid selection picks from the incorrect dominant cluster---since \name{} only selects
among the sampled candidates and cannot produce a successful chunk
if none would succeed; we treat this as an assumption and
discuss it in Section~\ref{sec:limitations}.

\begin{table}[!htbp]
  \centering
  \small
  \setlength{\tabcolsep}{6pt}
  \renewcommand{\arraystretch}{1.05}
  \caption{\textbf{Comparison with TACO on RoboTwin~2.0
    with~$\pi_{0.5}$.} Each task is evaluated 50 times.}
  \label{tab:taco-comparison}
  \begin{tabular}{lccc}
    \toprule
    \textbf{Task} & \textbf{Baseline ($K{=}1$)} & \textbf{TACO} & \textbf{\name{}} \\
    \midrule
    Adjust Bottle        & 88.0\% & \textbf{94.0\%} & 92.0\% \\
    Beat Block Hammer    & 68.0\% & 66.0\% & \textbf{70.0\%} \\
    Handover Block       & 24.0\% & 30.0\% & \textbf{34.0\%} \\
    Move Can Pot         & 42.0\% & \textbf{60.0}\% & 54.0\% \\
    Place Object Stand   & 68.0\% & \textbf{80.0}\% & 72.0\% \\
    \midrule
    Overall              & 58.0\% & \textbf{66.0\%} & 64.4\% \\
    \bottomrule
  \end{tabular}
  \vspace{-0.5em}
\end{table}

\paragraph{Comparison with TACO.}
On the five RoboTwin~2.0 tasks for which TACO~\citep{taco} has
released trained selectors, \name{} matches TACO at the macro
level ($64.4\%$ vs $66.0\%$, both at $K{=}16$ over a $58.0\%$
$K{=}1$ baseline; Table~\ref{tab:taco-comparison}). It does so
without TACO's per-task auxiliary training (hundreds of GPU hours), demonstrating that action-chunk geometry alone is a
powerful selection signal to match model based approaches.

\subsection{Ablations}
\label{sec:ablations}

We ablate three components of \name{}'s selector: the distance
function, the number of samples $K$, and the number of clusters
$C$. To control evaluation cost we run ablations on the five
vla-eval pairs in Table~\ref{tab:main-results}; each ablation
varies one component while holding the rest at their defaults (L2
distance, $K$ from Section~\ref{sec:latency-overhead}, $C{=}2$). Table~\ref{tab:ablations} reports the
resulting per-pair improvement over the $K{=}1$ baseline.

\begin{table}[!htbp]
  \centering
  \small
  \setlength{\tabcolsep}{6pt}
  \renewcommand{\arraystretch}{1.05}
  \caption{\textbf{Ablations on \name{}'s selector.} Each cell is
    the absolute (percentage-point) improvement over the $K{=}1$
    baseline. \emph{Default}: L2 distance, $C{=}2$, and $K$ chosen
    as in Section~\ref{sec:latency-overhead} ($K{=}4$ for
    GR00T~N1.6 and X-VLA, $K{=}16$ otherwise). The ablation
    columns vary one component at a time. Bold marks the largest
    gain per column.}
  \label{tab:ablations}
  \begin{tabular}{llcccc}
    \toprule
    \textbf{Workload} & \textbf{Model} & \textbf{Default} & \textbf{Cosine} & \textbf{$2K$} & \textbf{$C{=}4$} \\
    \midrule
    LIBERO                      & $\pi_{0.5}$  & \green{$+1.0$}            & \green{$+0.3$}           & \green{$\mathbf{+1.7}$}            & \green{$+1.1$} \\
    SimplerEnv -- WidowX        & GR00T~N1.6   & \green{$+13.3$}  & \green{$+0.5$}           & \green{$\mathbf{+14.2}$}  & \green{$+8.3$} \\
    SimplerEnv -- WidowX        & X-VLA        & \green{$\mathbf{+3.1}$}            & \green{$+1.7$}           & \green{$+2.7$}            & \green{$+1.4$} \\
    SimplerEnv -- WidowX        & StarVLA      & \green{$+6.9$}            & \green{$+3.1$}           & \green{$\mathbf{+7.5}$}            & \green{$+5.9$} \\
    SimplerEnv -- Google Robot  & GR00T~N1.6   & \green{$\mathbf{+7.3}$}            & \green{$+5.0$}  & \green{$+6.8$}            & \green{$+6.2$} \\
    \bottomrule
  \end{tabular}
\end{table}

\paragraph{L2 vs cosine.}
Cosine distance reduces the gain on every pair, most dramatically
on GR00T~N1.6/WidowX ($+0.5$ vs $+13.3$ for L2). Action chunks are
bounded physical commands whose magnitude is meaningful, so a
magnitude-blind metric loses the structure cluster-medoid relies on
(Section~\ref{sec:candidate-selection}).

\paragraph{Number of samples: $K$ vs $2K$.}
Doubling $K$ produces only small fluctuations around the default, indicating that the chosen $K$ is already near the
saturation point of mode estimation. $K$ therefore serves as a tunable knob for trading off latency and
accuracy: our chosen values sit near the accuracy saturation point
while preserving the latency, and operators can dial $K$ up
for higher and more stable accuracy with higher latency overheads.

\paragraph{Number of clusters: $C{=}2$ vs $C{=}4$.}
Increasing $C$ from $2$ to $4$ produces comparable or smaller gains on every pair. The default $C{=}2$ matches the
structure we observe in our workloads often have a single dominant mode plus scattered failure samples. Therefore, additional clusters
risk splitting the dominant mode itself.

% !TeX root = ./paper.tex

\section{Limitations}
\label{sec:limitations}

Like other existing test-time scaling methods for both language models and physical AI models, \name{} assumes that the
underlying model is properly trained and samples useful candidate
trajectories. It can choose a better action chunk among the sampled
candidates, but it cannot produce a successful action if none of the
sampled candidates would succeed, or if the model is confidently
wrong on the majority of samples---in that case cluster-medoid
selection picks from the incorrect dominant cluster and may even
amplify the error rather than suppress it. Additionally, our latency measurements use batch size one, a common robot-deployment
setting with a dedicated GPU. When the same accelerator is shared across multiple robots and the effective batch size is already large, the available headroom for additional parallel samples may shrink; in such cases, $K$ should be chosen by profiling the deployment workload.

\section{Conclusion}
\label{sec:conclusion}

Diffusion- and flow-matching physical AI models are powerful, but their
standard deployment loop still commits to one stochastic action chunk at
each round. \name{} addresses this brittleness with a simple
inference-time wrapper: sample several chunks from the same context,
use the geometry of continuous action space to find the dominant
consensus mode, and execute that mode's medoid as a real model output.
This enables judge-free self-consistency for action
generation. Two properties make \name{} effective and practical. First, action
chunks are compact, so parallel sampling can reuse the shared model
context and add little end-to-end latency. Second, the geometry of
output action chunks provides a natural signal for model confidence. Across VLA
and WAM testbeds, \name{} improves task success over single-trajectory sampling by up to 13.3\%, while remaining on par with model-based test-time scaling approaches. More broadly, \name{} shows that test-time
scaling for physical AI can be grounded with the geometry structure of the model's own output.

% !TeX root = ./paper.tex

\bibliographystyle{plainnat}
\bibliography{reference}

\appendix
% !TeX root = ./paper.tex

\section{Technical Appendices and Supplementary Material}
\label{sec:appendix}

\subsection{Implementation: \name{} as an Inference-Time Wrapper}
\label{app:implementation}

\name{} is implemented as a lightweight Python wrapper that
intercepts each action-generation call of an existing policy and
replaces it with $K$-sample parallel inference plus cluster-medoid
selection. The wrapper does not modify model weights, retrain any
component, or change the downstream control loop. Across the three
testbeds we evaluate on (LeRobot, vla-eval, Fast-WAM), the
integration follows the same three-step pattern at each round:
\begin{enumerate}
  \item \textbf{Shared context expansion.} The current observations
  and task instruction are encoded once. The resulting prefix
  tensors (VLM features, transformer KV cache, or world embedding)
  are given a candidate dimension and expanded to width $K$ via
  \texttt{torch.expand}, so that the $K$ candidates share the same
  underlying storage rather than materializing $K$ copies of the
  context.
  \item \textbf{Batched candidate sampling.} We draw $K$ independent
  noise tensors and run the model's denoising or flow-matching loop
  as a single batched forward pass, producing $K$ candidate action
  chunks $\{a_i\}_{i=1}^{K}$.
  \item \textbf{Cluster-medoid selection.} Each candidate is
  flattened, the $K\times K$ pairwise L2 distance matrix is computed
  in one batched tensor operation, and the threshold-guarded
  cluster-medoid procedure of Section~\ref{sec:candidate-selection}
  returns a single action chunk $a^\star$, which is handed to the
  existing controller in the original output format.
\end{enumerate}

\paragraph{LeRobot.}
For policies using the LeRobot inference loop ($\pi_{0.5}$ and
SmolVLA in our evaluation), we subclass each policy and override its
action-generation method. The override replaces the original noise
sampling and denoising/flow-matching steps with their batched
($K$-wide) versions, calls into the shared selector, and returns a
single chunk in the original output format. The downstream LeRobot
controller sees no change.

\paragraph{vla-eval.}
For policies served from vla-eval benchmark Docker images (GR00T
N1.6, X-VLA, and StarVLA in our evaluation), we wrap the inference
module inside each image. The wrapper reuses the image's existing
model loading, observation pre-processing, and action
post-processing, and intercepts only the function that maps an
encoded observation context to an action chunk. This keeps the
simulator, dependencies, and hardware setup byte-identical to the
reproduced baseline.

\paragraph{Fast-WAM.}
For the world-action model setting (Fast-WAM in our evaluation), we
wrap the per-round generate step. The video generation backbone runs
once per round to produce the shared world embedding; we expand the
embedding to a $K$-batch and run the action decoder's denoising loop
as a single batched pass. The selector runs identically to the VLA
case.

\paragraph{Selector implementation.}
The cluster-medoid selector is implemented in PyTorch as a small set
of batched on-GPU operations: a $K\times K$ pairwise L2 distance
matrix, a global-medoid argmin, the threshold-$\tau$ unimodality
guard (with $\tau = 0.3$ as the default), a small k-means loop
($C{=}2$, up to $10$ iterations) over the flattened action chunks,
and a within-cluster medoid argmin. All steps reuse the precomputed
distance matrix; the entire selection adds at most a few hundred
microseconds per round at the $K$ values used in our evaluation.

Listing~\ref{lst:wrapper} sketches the wrapper that intercepts an
existing policy's action-generation call, and
Listing~\ref{lst:selector} shows the corresponding cluster-medoid
selector.

\begin{lstlisting}[style=keystonepy,
                   caption={\textbf{Wrapper integration pattern.}
                            \name{} subclasses an existing policy
                            and replaces the per-round action call
                            with shared-context expansion, batched
                            $K$-sample generation, and cluster-medoid
                            selection.},
                   label={lst:wrapper}]
class KeyStonePolicy(BasePolicy):
    def __init__(self, base_policy, K=16, C=2, tau=0.3):
        self.base = base_policy
        self.K, self.C, self.tau = K, C, tau

    def predict_action(self, observation):
        # 1. Encode shared context once (VLM features, KV cache,
        #    or world embedding -- depends on the base policy).
        context = self.base.encode(observation)

        # 2. Expand to a K-batch *without* materializing K copies.
        context_K = expand_to_K(context, self.K)  # uses torch.expand

        # 3. Run K denoising or flow chains as one batched forward.
        noise = torch.randn(self.K, *self.base.noise_shape,
                            device=context.device)
        candidates = self.base.sample_actions(context_K, noise)
        # candidates: [K, T, action_dim]

        # 4. Cluster-medoid selection -- returns one real chunk.
        return cluster_medoid_select(candidates,
                                     C=self.C, tau=self.tau)
\end{lstlisting}

\begin{lstlisting}[style=keystonepy,
                   caption={\textbf{Cluster-medoid selector.}
                            All steps are batched on-GPU and reuse
                            the same pairwise distance matrix.
                            $\tau$ guards against forced clustering
                            of an already-unimodal candidate set
                            (Section~\ref{sec:candidate-selection}).},
                   label={lst:selector}]
def cluster_medoid_select(actions, C=2, tau=0.3, eps=1e-8):
    # actions: [K, T, action_dim]
    X = actions.flatten(start_dim=1)             # [K, D]
    K = X.shape[0]

    # Pairwise L2 distance matrix (reused throughout).
    dists = torch.cdist(X, X)                    # [K, K]

    # Global medoid (argmin of row sums of distances).
    m = dists.sum(dim=1).argmin()                # scalar

    # Unimodality guard: how far is the mean from the medoid,
    # normalized by the typical pairwise distance?
    iu = torch.triu_indices(K, K, offset=1)
    median_d = dists[iu[0], iu[1]].median()
    s = (X.mean(dim=0) - X[m]).norm() / (median_d + eps)
    if s < tau:
        return actions[m]                         # already unimodal

    # Otherwise: k-means over K compact vectors, take the
    # largest cluster's medoid.
    z = kmeans(X, C=C, max_iter=10)              # [K] cluster ids
    c_star = torch.bincount(z).argmax()
    mask = (z == c_star)
    sub_dists = dists[mask][:, mask]
    i_local = sub_dists.sum(dim=1).argmin()
    i_star = mask.nonzero(as_tuple=True)[0][i_local]
    return actions[i_star]
\end{lstlisting}

\end{document}